# Distributed Collision-free Protocol for AGVs in Industrial Environments

Dario Marino, Adriano Fagiolini, Lucia Pallottino

*Abstract*—In this paper, we propose a decentralized coordination algorithm for safe and efficient management of a group of mobile robots following predefined paths in a dynamic industrial environment. The proposed algorithm is based on a shared resources protocol and a replanning strategy. It is proved to guarantee ordered traffic flows avoiding collisions, deadlocks (stall situations) and livelock (agents move without reaching final destinations). Mutual access to resources has been proved for the proposed approach while condition on the maximum number of AGVs is given to ensure the absence of deadlocks during system evolutions. Finally conditions to verify a local livelocks will also be proposed. In consistency with the model of distributed robotic systems (DRS), no centralized mechanism, synchronized clock, shared memory or ground support is needed. A local inter-robot communication, based on sign-boards, is considered among a small number of spatially adjacent robotic units.

## NOTES TO PRACTITIONERS

With the proposed approach Autonomous Guided Vehicles will be able to move safely (without collisions) in an industrial environment. Contrarily to what is usually done in plants there is not a central unit that coordinates the AGVs. The coordination is based on a set of rules such as those in urban streets and AGVs apply the same set of rules taking decisions of motion based on the positions and other information of neighbouring AGVs. Distributing the decision from a central authority to the vehicles make the system more robust to possible failures. If the central authority fails the entire system is out of control. In a distributed approach the failure would involve a single AGV and it would affect only few other vehicles.

Moreover, with the proposed approach we also try to solve the problem of possible stall configurations from which vehicles are not able to move or configuration from which agents can move without reaching their final destinations.

## I. INTRODUCTION

Conflict resolution problem in multi-vehicle systems has received extensive attention in the literature. For industrial applications, safety (in terms of collision avoidance), robustness and efficiency of operations in material handling systems

The Authors are with the Interdepartmental Research Center "E. Piaggio", Faculty of Engineering - University of Pisa, e-mail: {d.marino, a.fagiolini, l.pallottino,}@centropiaggio.unipi.it

This work has been partially supported by CONET, the Cooperating Objects Network of Excellence, funded by the European Commission under FP7 with contract number FP7-2007-2-224053 and Contract IST 224428 (2008) (STREP) "CHAT - Control of Heterogeneous Automation Systems: Technologies for scalability, reconfigurability and security"and the FP7-ICT-2009-2130 "PLANET".

are critical. Moreover, stall situations resolution and fluent navigation of the agents should also be guaranteed. Stall situations occur when agents are not able to move from the particular configuration (i.e. deadlock) or are constrained to move along a finite number of paths without reaching the final destination (i.e. livelock).

In existing literature, coordination of multi-vehicle systems has been exhaustively discussed, especially for Autonomous Guided Vehicles (AGVs). Most of these works may be classified in two categories, centralized or decentralized control policies. The centralized control policies ensure optimal control for agents navigation and prevent collisions and stall situations assigning to agents a priori determined paths. Even if this approach is very powerful for what concerns robustness, safety and collision avoidance, it is strictly constrained by the computational time requested for real-time motion of the agents, that increases with the number of the agents involved. Moreover, the faster must be executed the coordination algorithm, the more expensive must be the hardware used for computing the paths, and this is strictly connected to the cost of the hardware used to run the coordination protocol. Another disadvantage of centralized control policies is that if the central control unit fails, the whole system is out of control.

Many works are focusing on the important aspects of collision avoidance and deadlock avoidance, see e.g. [1], [2], [3], [4] In many cases ([5], [6]) Petri's net based control policies are adopted to avoid deadlocks through a path replanning or using a master-slave approach ([7]). Other strategies ([8], [9]) avoid deadlocks trying to detect a cyclic-waiting situation, using graph theory for planning paths such that deadlock is a-priori avoided ([10]) or using a matrix-based deadlock detection algorithm ([11]). Decentralized control policies resolve the issue of high computational costs, considering the problem divided in two phases: first, for each agent an optimum path is defined according to some cost index; then each agent and its neighbors resolve locally and by themselves conflicts that could arise, according to a shared coordination policy. There are many decentralized solutions in the literature: hybrid centralized-decentralized control ([12], [13]); decentralized shared control policy ([14]); discretization of the operative space in finite spatial resources ([15], [16]), which are contended between agents using a distributed mutual exclusion policy ([18], [20], [21], [24]); optimal path assignment and a replanning in case of deadlocks ([22]).

With respect to previous works, this paper presents: a decentralized control mechanism that make the whole system more scalable, robust and less computationally expensive, so that high performance hardware is not required; a smart replanning algorithm that ensures a safe and fluid navigation of the agents, avoiding deadlocks; a rule based protocol for coordination shared among the agents; a simple and easy to implement communication system based on virtual sign–boards, that causes less overloads on the network with respect to message passing and that is more efficient for communication between agents. All agents operate autonomously based on local communication with other agents inside a defined communication radius: no centralized mechanism, synchronized clock, shared memory or ground support is needed.

## II. PROBLEM DEFINITION

### A. Industrial environment

The considered environment is structured and can be represented, based on spatial discretization, by a finite set of resources. A graph $G=(V,E)$ is a natural representation of the discretized environment, where $V = \{n_1, \ldots, n_\varepsilon\}$ is the set of nodes and $E \in V \times V$, is the set of arcs that represents connections between nodes.

Each node represents an area of the environment and is characterized by an *ID*, that uniquely identifies the node, and by the position $(x, y)$ of a given point of the area (e.g. the center) in the global reference system.

A simple example of an industrial layout discretized in resources is reported in Fig. 1.

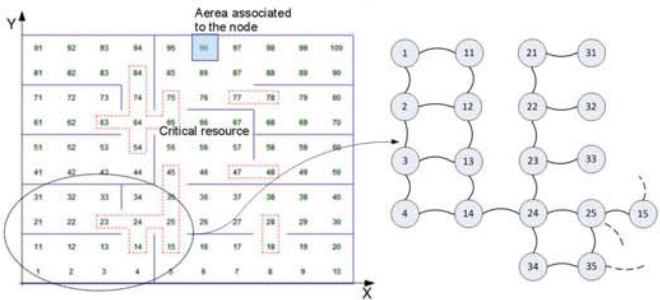

Fig. 1. Example of discretized industrial layout and part of the associated graph

Typically, an industrial environment consists in wide (rooms) and critical regions (such as doors, alternating one-lane corridors, etc.) where the coordination of motion is more challenging.

Each region can be represented by one or more nodes. A room may consists in $N_R$ (non critical) nodes and $N_C$ critical nodes, e.g. from Fig. 1 the bottom left room consists in $N_R = 7$ nodes and $N_C = 1$ critical node. We denote with $R_i \subset V$ the $i-th$ room and with $C_i \subset V$ the $i-th$ critical area, e.g. from Fig. 1 the bottom left room consists in $\{1, 2, 3, 4, 11, 11, 13, 14\}$ and the bottom left critical area consists in $\{14, 15, 23, 24, 25, 35, 45\}$.

### B. Shared resources

Resources can be managed at node level, as in the work [16] where only the first desired node (*Micro Resource*) is contended between agents. This kind of approach is clearly not efficient in terms of optimizing the traffic flow, and it is not able to avoid or prevent system deadlocks. On the other hand, resources can be managed as sets of nodes (*Macro Resources*), as e.g. in [17], where a set of desired nodes is contended and managed in order to avoid deadlocks a priori.

In our approach, the coordination system of each active agent within the logistic area will manage the access to shared resources at two levels:
- *MACRO*[1] *LEVEL*: each agent competes for obtaining the right to access to its macro resource;
- *MICRO*[1] *LEVEL*: once it has obtained access to a macro resource, each agent competes to use its individual parts, i.e. the micro resources.

The resources management at macro level will be used to coordinate agents in an efficient way and to ensure the absence of system deadlocks, whereas the resources management at the micro level will be used to avoid collisions.

More formally, every single node of $A_i$'s path $p_{A_i}$ is a macro resource. Hence, for each agent, the set of micro resources $mR_{A_i}$ represents the set of path nodes:

$$mR_{A_i} = p_{A_i} = \{n_{i,s}, \ldots, n_{i,f}\} \tag{1}$$

*Definition 1:* The micro resource $n_\nu \in mR_{A_i}$ of $A_i$ is a *shared micro resource* with $A_j$ if $n_\nu \in mR_{A_i} \cap mR_{A_j}$.

*Definition 2:* A *macro resource* $MR_{i,j} = \{n_{i,1}, n_{i,2}, \ldots, n_{i,d}\} \subset p_{A_i}$, of agent $A_i$, is an ordered sequence of future consecutive shared micro resources of its path. Let $MR_{A_i} = \{MR_{i,1}, \ldots, MR_{i,r}\}$ be the set of $r$ macro resources of agent $A_i$.

*Definition 3:* $A_i$ and $A_j$ *share* a macro resource $MR_{i,k}$ if it contains a micro resource $n_\nu \in mR_{A_i} \cap mR_{A_j}$.

Notice that a single macro resource can be shared (also partially) among multiple agents (see Fig. 2).

*Definition 4:* Given the current and the next desired nodes $n_{i,c}$ and $n_{i,c+1}$, the agent configuration is $l_{A_i} = \{n_{i,c}, MR_{i,\kappa}\}$, if $n_{i,c+1}$ belongs to a shared resource $MR_{i,\kappa}$ in $MR_{A_i}$ otherwise $l_{A_i} = \{n_{i,c}, n_{i,c+1}\}$.

An example of micro and macro resources and configurations is reported in Fig. 3. In this case, for agent $A_3$ we have:

$$\begin{cases} p_{A_3} = mR_{A_3} = \{n_{3,s}, \ldots, 6, 5, 4, 3, 9, \ldots, n_{3,f}\} \\ MR_{A_3} = \{\ldots, MR_{3,i}, MR_{3,i+1}, \ldots\} \rightarrow \begin{cases} MR_{3,i} = \{5\} \\ MR_{3,i+1} = \{3\} \end{cases} \\ l_{A_3} = \{n_{3,c}, MR_{3,i}\} = \{6, 5\} \end{cases}$$

[1]In the paper all variables with subscript $M$ and $m$ will refer to MACRO and MICRO respectively.

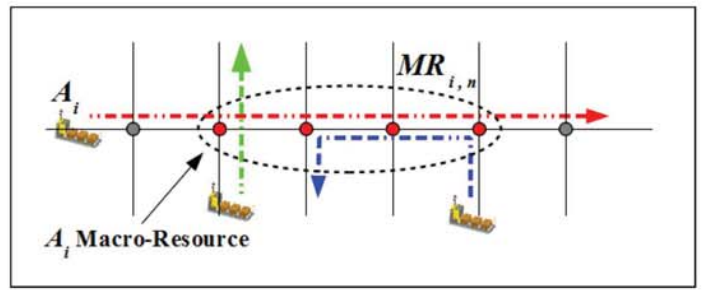

Fig. 2. Example of a macro resource shared partially with multiple AGV

For a generic AGV $A_i$, the desired path $p_{A_i}$ is an ordered sequence of nodes that corresponds to a path of $G$, e.g. $p_{A_i} = \{\eta_{i,s}, \ldots, \eta_{i,f}\}$. Let $l_{A_i} = \{\eta_{i,c}, \ldots, \eta_{i,r}\} \subset p_{A_i}$ be the $A_i$ configuration that represents the current node $\eta_{i,c}$ and the following desired nodes, see Definition 4.

In a traffic network there are typically three typologies of encounters between pairs of vehicles characterized by the resources that needed to be shared (see Fig. 4), [17].

Let $\eta$ be a sequence of ordered nodes, e.g. $\eta = \{\eta_1, \ldots, \eta_m\}$. We denote with $\overline{\eta} = \{\eta_m, \ldots, \eta_1\}$ the reverse sequence. Furthermore, let $\#\eta = m$ be the number of nodes in $\eta$.

The three encounter typologies (or sub-sequences of nodes) between $A_i$ and $A_j$ are characterized as follows:

(T1) *CROSSROAD*: if the only sub-sequence of nodes in both $l_{A_i}$ and $l_{A_j}$ consists in a single node;

(T2) *FOLLOWER*: if the sub-sequences of nodes in both $l_{A_i}$ and $l_{A_j}$ consists in a sequence of at least two nodes;

(T3) *FRONTAL*: if the sub-sequences of nodes in both $l_{A_i}$ and $\overline{l}_{A_j}$ consists in a sequence of at least two nodes.

Notice that more complicated encounters consist of a combination of above mentioned typologies.

## C. AGVs' Dynamics

The generic configuration of each agent is defined by a vector $q_i = \{x_i, y_i\}$ containing its position in the global reference system:

$$\dot{q}_i = u_i = \begin{cases} \dot{x}_i = u_{i,x} \\ \dot{y}_i = u_{i,y} \end{cases} \quad (2)$$

where $u_{i,x}$ and $u_{i,y}$ represent forward speed components along the two axes of the global reference system, respectively.

## D. Communication model: Sign–board

Communication for motion coordination is allowed only between agents inside a given communication radius. In other words agents $i$ and $j$ communicate only if their Euclidean distance is less than a given value. In this case we refer to $i$ and $j$ as neighboring agents, i.e. $A_j \in neigh_{A_i}(t)$ and $A_i \in neigh_{A_j}(t)$. The communication model used in this paper is based on virtual sign–boards ([19]). Each robot updates dynamically its own sign–board $SB_i(t_k)$ and can have access to the sign–boards of neighboring agents. At each discrete execution time $t_k$ of the algorithm, the sign–board is characterized by the fields shown in the following table and each agent reads sign–boards of neighbouring robots, $lect_i(t_k) = \{SB_j(t_k) \mid A_j \in neigh_{A_i}(t_k)\}$.

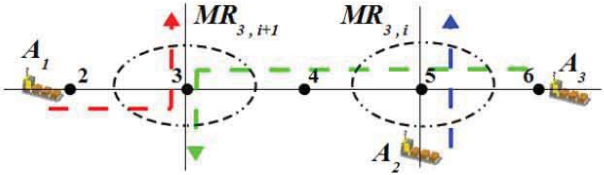

Fig. 3. Example of micro and macro resources shared with multiple AGV

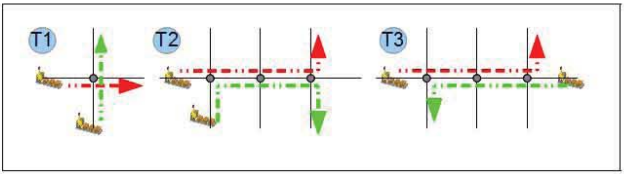

Fig. 4. Types of encounters: (T1) crossroad, (T2) follower, (T3) frontal

| Field | Description |
|---|---|
| ID | Identification Number $i$ |
| Pr | Priority value $Pr_i$ |
| Status | State value $\sigma_i$ |
| Vel | Speed value $u_i$ |
| Nodes(1:N) | Sequence of nodes representing $p_{A_i}$, $N$ is at most the total number of nodes in the layout |
| Curr | Current node $curr_i$ |
| Next | Next node in the path $p_{A_i}$, $next_i$ |
| Prev | Previous node in the path $p_{A_i}$, $prev_i$ |
| Timer | time spent waiting to access to a critical resource, when resource is free again competition on Timer value |

TABLE I
SIGN–BOARD FIELDS

## E. Priorities for coordination

A priority order based policy is used to manage motion coordination between agents. The order of access to a shared resource is given by the order of priorities of the agents involved in the competition. A higher priority implies that the agent has more chances to win a competition. If two or more agents have the same priority, ID is used for resolving competition.

For each iteration of the algorithm, agent $A_i$ reads sign–boards of neighboring robots in $neigh_{A_i}$, detects shared resources and determines the priorities of the agents to compete with.

Priority of each agent is static and can be assigned by the user at the beginning of the task. In particular cases the priority is modified during protocol execution. Indeed, when an agent occupies a critical resource, such as doors or corridor, its priority is set at the maximum value so that any other agent can't enter the same critical macro resource before the agent releases it. A dynamic priority scheme such as the one proposed in [24], can also be used. By taking into account other information as the mission priority level, AGVs battery level, etc., it is possible to extend the priority scheme to more complex behaviors.

## III. THE COLLISION-FREE MOTION PROTOCOL

According to the proposed model, based on the Distributed Robotic Systems (DRS) one, $n$ agents $A_i, \ldots, A_n$ interact with each other by following a shared motion cooperation protocol (see e.g. [23]). While the continuous dynamics of each vehicle is described by Eq. 2, agents' resource competition can be described as a Discrete Event System (DES), and indeed Finite State Machine (FSM), which is described below. Every agent $A_i$ runs an event-driven *cooperation manager* which ensures the correct execution of the shared common protocol.

The manager receives neighboring agents' sign–boards, $lect_i(t_k)$, and outputs the vehicle speed $u_i(t)$ and the updated sign–board $SB_i(t_k)$. Thus, the agent's dynamics evolves according to a hybrid dynamic $(\dot{q}_i(t), \sigma_i(t_{k+1}), SB_i(t_{k+1})) = H(q_i(t), \sigma_i(t_k), SB_i(t_k), lect_i(t_k))$, where $H$ depends on the set of variables and maps $\{q_i, \sigma_i, \zeta_i, SB_i, lect_i, f_i, g_i, s, \delta, r, d\}$ described below. A depiction of an agent's architecture is reported in Fig. 6. In the remainder of this section a description of the model is given.

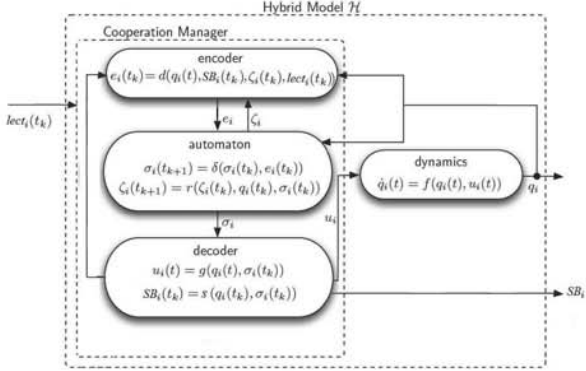

Fig. 6. Hybrid model describing the architecture of a generic agent.

At each execution time of the algorithm the state $\sigma_i(t_k)$ of an agent's cooperation manager takes values in the set

$$\Sigma = \{REQUEST, WAIT, MOVE, REPLAN\}.$$

where each value has the following meaning:

- REQUEST (Req): the agent competes to access to its next node. In this state, the sequence of operations is:
  1) acquisition of the neighbours sign–boards $lect_i(t_k)$;
  2) macro resources detection $MR_{A_i}$;
- MOVE (M): the agent move along its path with the maximum allowed speed.
- WAIT (W): the agent doesn't move and waits to access its next node.
- REPLAN (Rep): the agent replans its path up to the final node.

The discrete evolution of state $\sigma_i$ is given by

$$\begin{cases} \sigma_i(t_{k+1}) = \delta_i(\sigma_i(t_k), e_i(t_k)), \\ \sigma_i(0) = Req. \end{cases}$$

where $\delta_i$ is the automaton, or finite state machine, and where given the lecture $lect_i(t_k)$, events $e_{i,j}$ are generated by the *encoder* described next.

## A. Encoder

All events triggering a change from the current maneuver to a new one can be formalized as a set of binary variables, whose values depend on the configuration $q_i(t)$ of the agent, its sign–board $SB_i(t_k)$, and the sign–boards of its neighbors,

$lect_i(t_k)$. Each event is obtained as a logical sum of some sub–events $s_{i,m}$,

$$e_{i,l} = \bigwedge_{m \in S(event_l)} s_{i,m}$$

where S returns the set of indices of the sub–events composing a given event.

For the sake of clarity we do not report the events in details but we describe the switches between states referring to the sign–board variables in Table I. Switches and events that generate them are reported in Fig. 5.

Considering agent $A_i$ in $Req$ state, it first checks whether its next desired node $next_i$ do not belongs to a critical resource $C_k$. In this case, if there exists $A_j$ that occupies $next_i$ and it is arrived at its goal node (i.e. $curr_j = next_i = goal_j$) agent $A_i$ switch to the $Rep$ state (i.e. replans in order to find another path towards its goal). Otherwise, if $curr_j = next_i$ but $A_j$ is not at its goal (i.e. $curr_j \neq goal_j$) agent $A_i$ switch to the $W$ state in order to be able to request the same resource at next time step. In case that no agent is occupying node $next_i$ competition is possible. If there exists $A_j$ requesting access to $next_i$ with $Pr_i > Pr_j$, agent $A_i$ wins the competition and switch to the $M$ state. Otherwise, having lost the competition it switch on its internal timer $t_{rep}$ and switch to the $W$ state. When the internal timer becomes larger than a given threshold a switch from $W$ to the $Rep$ state occurs.

It from the first check on $next_i$ turns out that the node belongs to a critical resources $C_k$ the procedure is different. The resource $C_k$ is not shared the switch to the $M$ state occurs. Otherwise, let $A_j$ be the AGV that shares $C_k$ with $A_i$. If $A_j$ already occupies a node of $C_k$, $A_i$ state switches to $W$ (as described later, after this switch the TIMER variable of the sign–board is switched on). Otherwise (i.e. nodes of $C_k$ are not occupied), if for all agents sharing $C_k$ with $A_i$ the TIMER variables are equal to zero (i.e. are switched off and set to zero) the competition of the resource is done on the priorities. In other words, if $A_i$ has largest priority it switches to the $M$ state, otherwise it switches to $W$ (also in this case, after this switch the TIMER variable of the sign–board is switched on). In presence of at least one agents that shares $C_k$ with $A_i$ and positive TIMER variables (i.e. TIMER variable activated), the resource competition is done on the TIMER variables. Indeed, if agent $A_i$ has largest TIMER values (i.e. is waiting for the resource from the largest amount of time) win the competition and switch to the $M$ state (in this case the TIMER variable is switched off and set to zero). Otherwise, it switches to the $W$ state and the TIMER variable increases of 1.

Switches to $W$, $M$ and $Rep$ correspond to changes of the sign–board as determined by the *decoder* and described below. However, from $W$ and $Rep$, after the decoding procedure is finished, a switch to the $Req$ state occurs. Finally, from $M$ a switch to $Req$ occurs whenever the current node $curr_i$ is updated (the agent leaves the region associated to the node and enters the next one).

As a whole, the encoder can be seen as a logical map allowing computation of the events as follows:

$$e_i(t_k) = d(SB_i(t_k), lect_i(t_k), \zeta_i(t_k), q_i(t)).$$

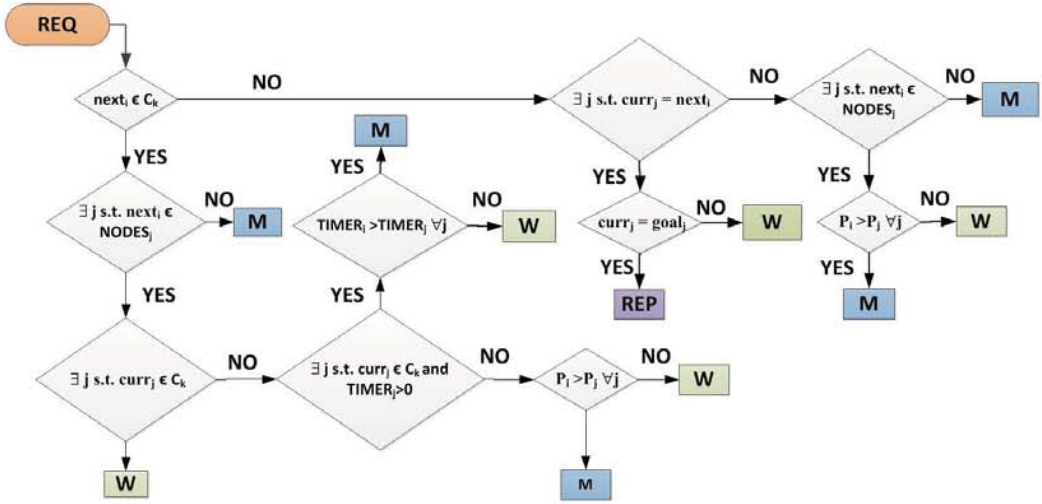

Fig. 5. The state flow representation from the $Rep$ state

### B. Decoder

The *decoder* is the application that produces the output of the hybrid system, i.e. AGV velocity $u_i(t)$ and the updated sign–board $SB_i(t_k)$, based on the current state $\sigma_i(t_k)$ and the current agent configuration $q_i(t)$. The two updating laws are

$$\begin{cases} u_i(t) = g_i(\sigma_i(t_k), q_i(t_k)), \\ SB_i(t_k) = s_i(\sigma_i(t_k), q_i(t_k)). \end{cases} \tag{3}$$

*1) Speed update:* The speed $u_i$ of each agent is updated according to its current state $\sigma_i(t_k)$ and its current configuration $q_i(t)$.

For the states $\sigma_i(t_k) \in \{Req, Rep, W\}$ the AGV velocity is set as follows

$$\begin{cases} u_i(Req) = u_i(t_{k-1}), \\ u_i(W) = [0, 0]^T, \\ u_i(Rep) = [0, 0]^T. \end{cases} \tag{4}$$

If the agent has won the competition for its next node, and hence its state is $M$, the velocity must be set taking into account the possibility that another AGV is still physically releasing the node. In this case the velocity is decreased from the maximum speed $u_{i,max}$ to a safe value $\bar{u}_i$, in order to avoid collisions. Referring to Fig. 7, let introduce some parameters to embed a node in the reference frame: let $u_{i,max}$ be the maximum speed, $D_{i,1}$ ($D_{i,2}$) the eucledean distance of previous (next) and current node, $\theta_{i,1}$ ($\theta_{i,2}$) the angle between segment $D_{i,1}$ ($D_{i,2}$) and horizontal (vertical) axis in the global reference system. The speed $\bar{u}_i$ is defined as follow:

$$\bar{u}_i = \begin{cases} u_k, & \text{if there exists an agent } A_k \text{ physically} \\ & \text{occupying the node won by } A_i \\ u_{i,max}, & \text{otherwise.} \end{cases}$$

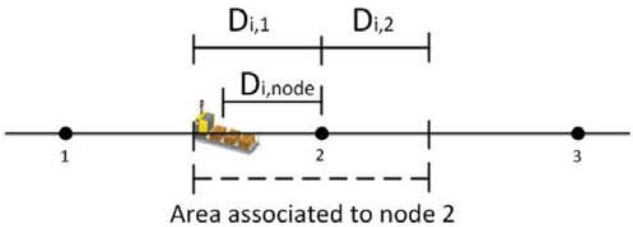

Fig. 7. Parameters for determining speed of the agents

Notice that $u_k$ is known because it is shown in the sign–board of agent $A_k$. Finally, the velocity in the $MOVE$ state is set

$$u_i(M) = \bar{u}_i \begin{pmatrix} \cos(\theta_{i,j}) \\ \sin(\theta_{i,j}) \end{pmatrix} \quad (x_i, y_i) \in D_{i,j} \tag{5}$$

So $A_i$ starts to move before the area associated to its next node is physically free, but with a speed that ensure collision avoidance, reducing idle time.

*2) Path replanning:* The *decoder* also provides the update of the sign–board $SB_i(t_k)$ of the agent $A_i$ at discrete time $t_k$ according its current state $\sigma_i(t_k)$ and current configuration configuration $q_i(t_k)$.

When a switch to the *REPLAN* state occurs, the original path of the agent is modified. Given the current node $curr_i$ of agent $A_i$, let $neigh(curr_i) = \{n_k \mid (curr_i, n_k) \in E\}$ be the set *neighbouring nodes* of $curr_i$, i.e. set of nodes connected with $curr_i$ through an arc of the graph. Given the lecture $lect_i(t_k)$, all neighbouring nodes currently occupied by other AGVs are marked as *occupied* and put in the list $O_i$ of occupied nodes. All other neighbouring nodes are marked as *free* and put in the list $F_i$ of free nodes.

Given the adjacency matrix $M \in \mathbf{R}^{N \times N}$ that describes the connections between nodes in the graph, where $m(i, j) = 1$ if $(n_i, n_j) \in E$ (i.e. the default weight associated to the arcs is 1), and given the lecture $lect_i(t_k)$, the adjacency matrix is temporarily modified. For each neighbouring node $n_i \in neigh(curr_i)$, if $n_i$ is a node of the path $p_{A_k}$ of agent $A_k$ all weights of the arcs associated to the sub-path of $p_{A_k}$ from $n_i$ to the last node of $p_{A_k}$ are increased. The increase for each arc is proportional to the length of the sub-path from the starting node of the first arc of the sub-path to the last node of $p_{A_k}$. In this way the weight is larger for arc closer to $n_i$ and decreases linearly towards the arc of $p_{A_k}$.

Once all nodes in $neigh(curr_i)$ have been analyzed a new adjacency matrix has been obtained and can be used to compute a minimum cost path (based on classical shortest path algorithms, see e.g.[25]) from $curr_i$ to the goal node of $A_i$.

For any $n_k \in neigh(curr_i) \cap F_i$ (i.e. free neighbouring nodes of $curr_i$) a minimum cost path $path_{free,k} = \{curr_i, n_k, \ldots, goal_i\}$ from $curr_i$ to $goal_i$ (through $n_k$) can be obtained. Given this set of new paths, the one at minimum cost will become the new path $p_{A_i} = min_j(path_{free,j})$ of agent $A_i$.

If $F_i = \emptyset$, for each neighbouring occupied node $n_j \in neigh(curr_i) \cap O_i$ a minimum cost path $path_{occ,j} = \{curr_i, n_j, \ldots, goal_i\}$ from $curr_i$ to $goal_i$ (through $n_j$) can be obtained. Given the set of minimum cost paths, the one at minimum cost will become the new path $p_{A_i} = min_j(path_{occ,j})$ of agent $A_i$.

The new path is finally updated in the sign–board of agent $A_i$. An example of path replanning is shown in Fig. 8.

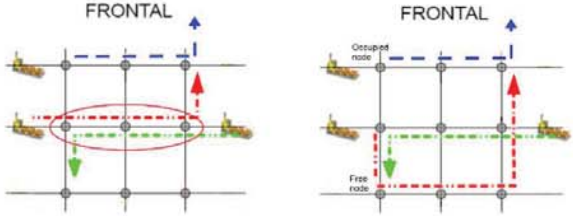

Fig. 8. Example of path replanning in a frontal encounter

*3) Sign–board update:* We now show how for each state $\sigma_i(t_k)$ of agent $A_i$, the sign–board $SB_i$ is updated. For the $REQUEST$ state the sign–board does not change except for the "Status" variable that is set to R, i.e. $Status = R$.

For the $WAIT$ state the "status" variable is changed into $W$ and the speed of the agent changes how it has been described previously in this section: $Vel = g_i(W, q_i(t))$ such that $u_i = 0$, refer to 3 and the speeds update laws. The TIMER variable update has been described in section III-A, it is activated in case of competition of a critical resource and is needed to give higher priority to agents that are waiting to access the resources for longer.

Similarly to the $WAIT$ state, in the "REPLAN" state the "status" variable is set to $Rep$ and $Vel = g_i(Rep, q_i(t))$ such that $u_i = 0$. In addiction, the new computed path must be updated in the sign–board, i.e. $Nodes(1:28) = p_{A_i}$, and the next desired node is set as the second node in $p_{A_i}$, i.e. $Suc = p_{A_i}[2]$. In this state the internal timer $t_r$, described in section III-A, is switched off (if on) and set to zero.

Finally in the $MOVE$ state we have $Status = M$ and $Vel = g_i(M, q_i(t))$. Let $\bar{t}_i$ the time instant in which $A_i$ physically enters into the area represented by its next node. If $t = \bar{t}_i$ then $Nodes(1, 28) = [Nodes(2:28), 0]$, $Prec = Act$, $Act = Nodes[1]$ and $Succ = Nodes[2]$, i.e. the list of nodes in the path is updated and the previous node released updating the new "current" and "next" nodes. If on, the TIMER variable is switched off and set to zero.

## IV. COLLISION AVOIDANCE

The most important goal of the coordination algorithm is to ensure collision avoidance between AGVs. This issue can be reached providing mutual exclusive access to a spatial resource whenever two or more agents compete for it. In the proposed algorithm, the main feature during the competition is the priority of the agents involved. When an AGV requests the access to its next resource, it checks the priority of the other robots interested to the same resource looking at their sign–boards and then, according to the shared coordination protocol, it compute its next state.

For the sake of simplicity, the coordination mechanism will be illustrated only in the case of two agents $A_i$ e $A_j$ competing for the same resource. Without loss of generality we can assume that $Pr_i > Pr_j$ and $u_i > u_j$. Demonstration is easily extendable to the generic case with $N$ agents.

Referring to Fig. 9, let introduce the following quantities:

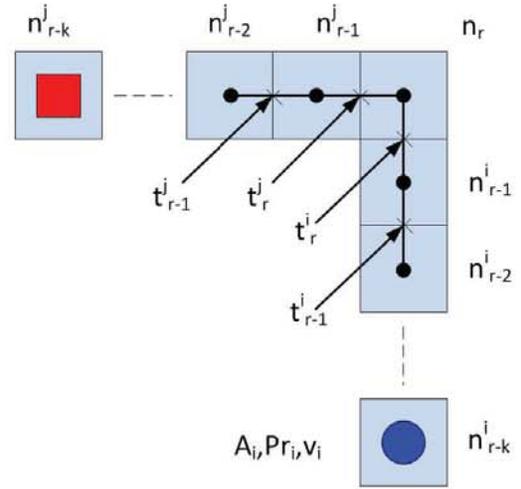

Fig. 9. Mutual exclusion quantities

- $n_r$: node requested both by $A_i$ and $A_j$;
- $n_{r-k}^i$: k-th node before $n_r$ in $p_{A_i}$, such that $path_i = \{n_{r-k}^i, \ldots, n_{r-1}^i, n_r\}$ ;
- $t_l^i$: time instant at which $A_i$ access to $n_l \in path_i$;
- $C_k$: k-th critical resource interested in the competition.

Agents are supposed to execute the algorithm with different sampling time and no synchronized clocks, so that during a competition requesting time $t_{R_i}$ and $t_{R_j}$ to the access $R$ may be different.

We introduce the constraint that, given the maximum speed of the agents, the sampling time of the algorithm is short enough to ensure at least two execution time of the algorithm for each spatial resources if the robot is moving at its maximum speed, before it enters into its next node. Let $d$ be the maximum dimension associated to the aerea of the nodes, $u_{i,max}$ is the maximum speed for agent $A_i$, $t'_{R_i}$ and $t''_{R_i}$ are instants at which agent $A_i$ request to access to its next node $R = next_i$ and $t_{next_i}$ the instant at which $A_i$ enters in node $next_i$.

The constraint is that

$$\forall A_i, T_i = \frac{u_{i,max}}{d/2} \implies \exists t'_{R_i}, t''_{R_i} < t_{next_i} \mid t''_{R_i} = t'_{R_i} + T_i.$$

Under this constraint we are able to ensure a mutual exclusive access to a shared resource.

*Theorem 1:* If the constraint on the sampling time for each agent is satisfied and if for $t^k_{n_{r-1}} < t < t^k_{n_r}$, $k = i, j$, AGVs are in $curr_i(t) = n^i_{r-1}$ and $curr_j(t) = n^j_{r-1}$, with $Pr_i > Pr_j$, then $\nexists t^* > t^i_{n_r}, t^j_{n_r} \mid curr_i(t^*) = n_r, curr_j(t^*) = n_r$.

*Proof:* Let $R = n_r$, given $curr_j(t^j_{R_j}) = n^j_{r-1}$, two cases are possible: $A_i$ is already in $curr_i(t''_{R_j}) = n_{r-1}$ or still in $curr_i(t''_{R_j}) = n_{r-2}$.

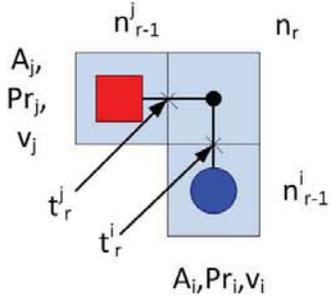

Fig. 10. Mutual exclusion: asynchronous case 1 of theorem 1

In the first case, referring to Fig. 10, we suppose by an absurd that there exists a time $t^* > t^i_{n_r}, t^j_{n_r}$ in which both agents access to the resource at the same time, so that $curr_i(t^*) = curr_j(t^*) = n_r$. Hence there exists a time $t_{R_j} < t^j_{n_r} < t^*$ in which $A_j$ requests and (according to rules in section III-A) is allowed to occupy $n_r$, so that $\sigma_j(t^+_{R_j}) = M$. But this would be true only if $Pr_j > Pr_i$ and this is absurd because we supposed $Pr_j < Pr_i$.

In the second case, referring to Fig. 11, we have to distinguish between two other cases:

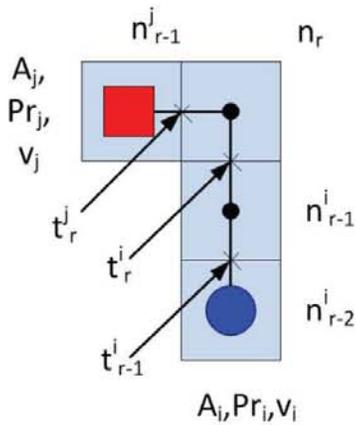

Fig. 11. Mutual exclusion: asynchronous case 2 of theorem 1

1) $A_j$ has access to $n_r$ before that agent $A_i$ has access to $n^i_{r-1}$, i.e. $t^j_{n_r} < t^i_{n_{r-1}}$;
2) $A_j$ has access to $n_r$ after that $A_i$ has access to $n^i_{r-1}$, i.e. $t^j_{n_r} > t^i_{n_{r-1}}$.

In the first case there exists a time $t'_{R_i} > t^i_{n_{r-1}} > t^j_{n_r}$ such that $A_i$ in $n^i_{r-1}$ sees (through $SB_j$) that $n_r$ is occupied by $A_j$. According to the shared coordination protocol, it will happen that $A_i$ is set to $WAIT$, i.e. $\sigma_i(t^k_{R_i}) = W$, and collision is avoided.

In the second case ($t^j_{n_r} > t^i_{n_{r-1}}$) let $t'_{R_i} > t^i_{n_{r-1}}$ be the instant at which $A_i$ requests access to $n_r$. If $t'_{R_i} > t^j_{n_r}$, the request is done after the node has been taken by $A_j$ hence the node is occupied and according to the coordination rules $A_i$ is set to the $WAIT$ status. Otherwise, if $t'_{R_i} < t^j_{n_r}$, according to the rules $A_i$ starts to move towards the node $n_r$ since $curr_j = n^j_{r-1}$ and for hypothesis $Pr_i > Pr_j$.

The distance $D^i_{n_r}$ of $A_i$ from $n_r$ is larger than the distance $D^j_{n_r}$ of $A_j$ from the same node, hence $A_j$ may reach $n_r$ prior than $A_i$. If this is the case, agent $A_i$ will recognize this event at its second lecture of the sign–boards, at the time $t'_{R_i} > t^j_{n_r}$, and in that moment it will stop, at a safe distance such that collision avoidance is ensured, thanks to the constraints on the algorithm execution time.

In this special case we have an inversion of the order of access to the resources, due to asynchrony but the coordination rules still ensure collision avoidance. Notice that increasing the frequency of the execution time, this particular situation has very little probabilities to be verified. ∎

## V. DEADLOCK/LIVELOCK AVOIDANCE

Another very important feature in coordination of AGVs, in order to improve efficiency of the material handling in an industrial layout, is to ensure the avoidance of situations of stall in which an agent continuously tries to access to a resource occupied by another robot (*deadlock*) or moving along a finite number of paths never reaching its goal node (*livelock*).

Given $N$ AGVs, we can define:

- $START = \{start_1, \ldots, start_N\}$ the set of non-equal starting nodes, i.e. $start_i \neq start_j, \forall i \neq j = 1, \ldots, N$;
- $GOAL = \{goal_1, \ldots, goal_N\}$ the set of non-equal goal nodes, i.e. $goal_i \neq goal_j, \forall i \neq j = 1, \ldots, N$, and such that for any of the $N_c$ critical resources $C_k$, $\forall i = 1, \ldots, N, \forall k = 1, \ldots, N_c, goal_i \notin C_k$, e.g. if an agent has as a goal a node representing a door could prevent the passage of other AGVs.
- Let $m_s = \min_i \#R_i$, i.e. the smallest dimension of the rooms $R_i$ of the industrial layout.

Given the sets *START* and *GOAL*, given $m_s$ and the set of neighbouring nodes $neigh(n_i)$, an upper bound on the number of AGVs that can be used in the environment to prevent deadlock situations can be determined.

*Remark 1:* Given the sets *START* and *GOAL* defined above, considering the worst case in which every goal node is in the room with smallest number $m_s$ of nodes, i.e. $\forall i, goal_i \in R_k$ with $\#R_k = m_s$, an upper bound on the number of agents that can be used in the layout is $N \leq m_s - 1$. Indeed, if $N > m_s - 1$, the set $GOAL$ of goal nodes has dimension $dim(GOAL) > m_s - 1$ and in the worst case there are more than $m_s - 1$ goal nodes in a room with $m_s$ nodes hence at least two AGV share a common goal node in contrast with the definition of the $GOAL$ set.

Every agent knows only a partial and uncomplete information of the state of the whole system, due to the decentralized nature of the coordination algorithm proposed, because it receives information only on its neighbouring agents, according to their real position in the environment and to the communication radius of the wireless system they have on board. For this reasons its impossible to detect global deadlock or livelock configurations.

However, every AGV can detect a local deadlock or livelock situation, according to the partial information owned, and can use this informatio to prevent the stall. A complete characterization of global deadlock or livelock configurations can be done only in a centralized framework.

### A. Local deadlock avoidance

We are interested in defining an operative procedure to locally detect if an agent $A_i$ is in a local deadlock (formally $A_i \in DEADLOCK$).

Let $p = \lfloor \frac{R}{d} \rfloor$ (the floor value of the ratio) where $R$ is the communication radius between agents and $d$ the maximum distance between consecutive nodes in the plant layout. Let $n_0 = curr_i$ the current node for agent $A_i$ and consider three sets of nodes $G$, $N_O$ and $F$, initialized to empty sets.

Let $N_0 = neigh(n_0) = \{n_1, \ldots, n_g\}$ be the set of neighbouring nodes of $n_0$. If there exists a node $n_i \in N_0$ that is free (i.e. in $O_{n_0}$) or not requested by another agent with higher priority, $A_i$ is not in a deadlock since it can move in $n_i$. For any node $n_i \in N_0$ that is occupied we proceed as follows. If there exists an agent $A_p$ that is arrived to its goal node, $curr_p = goal_p = n_i$, we put $n_i$ in $G$. If $n_i$ is currently occupied by an agent not yet arrived to its goal node $n_i = curr_i \neq goal_p$ we define a *first depth level* that is the set $N_1(n_{0,i}) = (neigh(n_i) \setminus \{n_0\})$ of neighbouring nodes of $n_i$, except $n_0$.

A family of first depth level sets $\aleph' = \{N_1(n_{0,1}), \ldots, N_1(n_{0,p}) | n_{0,j} \in N_0, j = 1, \ldots, p\}$ has then been obtained. For any $N_1(n_{0,j})$ with $n_{0,j} \in N_0$ we proceed as it has been done for $N_0$ with $n_{0,j}$ playing the role of $n_0$. Indee, for any $n_h \in N_1(n_{0,j})$, if $n_h$ is occupied by an agent $A_q$ that is arrived at its destination, i.e. $curr_q = goal_q = n_{hj}$, we put $n_h$ in $G$. Otherwise, if $N_1(n_{0,j}) \subset G$, then the node $n_{0,j} \in N_0$ is completely surrounded by nodes occupied by agents arrived at destination (except $n_0$) and we put $n_i$ in $N_O$.

If there exist $n_{0,j} \in N_0$ and $n_k \in N_1(n_{0,j})$ such that $n_k \notin G \cup N_O$, i.e. the node is not occupied or completely surrounded by occupied nodes except for $n_{0,j}$, the node is free and the procedure stops since the agent currently in $n_{0,j}$ can move in $n_k$ and $A_i$ in $n_{0,j}$ and there is no deadlock. Otherwise, for any pairs $n_{0,j}$, $n_k$ such that $n_k$ is occupied by an agent that is not yet arrived at its destination, we define a *second depth level* $N_2(n_{0,j,k}) = neigh(n_k) \setminus \{n_{0,j}\}$.

The procedure continues until the maximum allowed depth $p$ is reached, and the family $\aleph^{p-1}$ is defined, or until a deadlock is avoided in a smaller depth level $f < p - 1$.

Given $\aleph^f$ we put all free nodes in it in $F$. If $F$ is not empty there exists a displacement of $f$ AGVs that allows $A_i$ to move, hence no deadlock occurs.

An example of this procedure is given in Fig. 12, where in the pentagon node is the agent $A_i$, in red star nodes agents at their destinations, in light blue circles agents yet not arrived and in green triangle the free nodes:

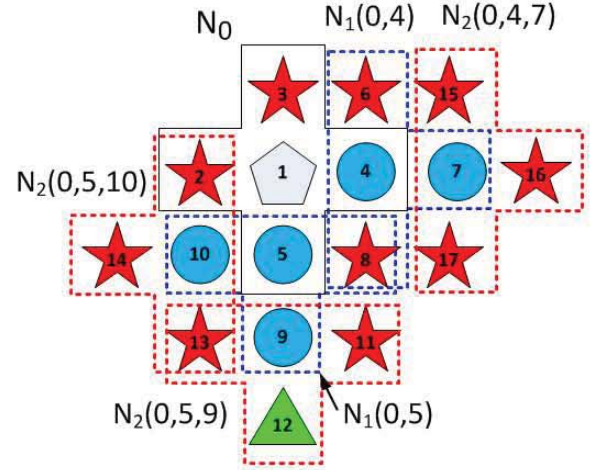

Fig. 12. Example of local deadlock identification procedure

- $R = 3, d = 1 \rightarrow p = 3$;
- $n_0 = curr_i = 1$;
- $N_0 = \{2, 3, 4, 5\}$;
- $F = \{2, 3, 6, 8, 11, 13, 14, 15, 16, 17\}$;
- $N_1(0, 5) = \{9, 10, 8\}$;
- $N_1(0, 4) = \{6, 8, 7\}$;
- $\aleph' = \{N_{0,5}, N_{0,4}\}$
- $N_O = \{7, 10\}$
- $N_2(0, 5, 10) = \{2, 14, 13\}$;
- $N_2(0, 5, 9) = \{11, 12, 13\}$;
- $N_2(0, 4, 7) = \{15, 16, 17\}$;
- $\aleph'' = \{N_{0,5,10}, N_{0,5,9}, N_{0,4,7}\} \equiv \aleph^f, f = p - 1 = 2$;
- $F = \{12\} \in \aleph_f$.

In this case $F \neq \emptyset$ hence a local deadlock does not occur.

*Theorem 2:* If $N \leq m_s - 1$, i.e. the constraint on the maximum number of agent is satisfied, and given the maximum depth $p$, if there exists at least one free node in $F$, then agent $A_i$ is not in a local deadlock situation according to the local information that it knows inside a finite number of steps $p - 1$, i.e. $A_i \notin DEADLOCK_{p-1}$.

*Proof:* Suppose by an absurd that hypothesis of the theorem are satisfied but agent $A_i$ is in deadlock. Then, there are no agents that can move in the next $p_1$ successive nodes. If so, for construction of the family of sets of depth level $p$, there are no free nodes in $F$. Hence the thesis. ∎

In the precedent example, given $p = 3$, there exists a free node $F = \{12\} \in \aleph^f, f = p - 1 = 2$ that the agents surrounding $A_i$ can use to move, hence $A_i$ is not in a deadlock.

## B. Local livelock avoidance

An agent is in *livelock* (formally $A_i \in "LIVELOCK"$) if it is blocked on a finite set of paths. In this case the agent can locally move switching between those paths, but it will never reach its final node. We can define a procedure that gives conditions that if satisfied ensure livelock avoidance.

With the same procedure described in the local deadlock avoidance case we are able to determine a set of paths $P_{safe}$ from the node $n_0 = curr_i$ for any $A_i$. When the procedure stops at level $p$ a set of paths of length shorter than $p$ has been determined if they exists.

An example of the procedure run by agent $A_i$ is shown in Fig. 13, where in green is shown $A_i$ and in red agents still arrived. The following variables are defined:

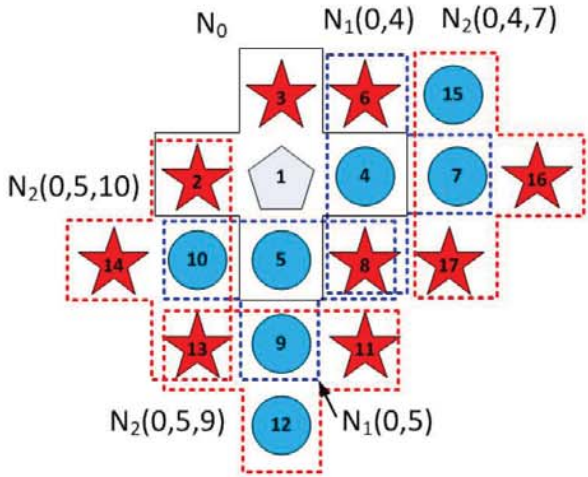

Fig. 13. Example of local livelock detection

- $R = 3, d = 1 \rightarrow p = 3$;
- $n_0 = curr_i = 1$;
- $N_0 = \{2, 3, 4, 5\}$;
- $F = \{2, 3, 6, 8, 11, 13, 14, 16, 17\}$;
- $N_1(0, 5) = \{9, 10, 8\}$;
- $N_1(0, 4) = \{6, 8, 7\}$;
- $\aleph' = \{N_5^0, N_4^0\}$;
- $N_2(0, 5, 10) = \{2, 14, 13\}$;
- $N_2(0, 5, 9) = \{11, 12, 13\}$;
- $N_2(0, 4, 7) = \{15, 16, 17\}$;
- $\aleph'' = \{(N_{10}^5)^0, (N_9^5)^0; (N_7^4)^0\}$;
- $N_O = \{10\}$;
- $P_{safe} = \{p'_{A_i}, p''_{A_i}\}; p'_{A_i} = \{1, 4, 7\}, p''_{A_i} = \{1, 5, 9\}$.

*Theorem 3:* If $N \leq m_s - 1$, i.e. the constraint on the maximum number of agent is satisfied, and given the maximum depth $p$, if there exists at least one path in $P_{safe}$ of length $p - 1$ and $A_i$ is not in a local livelock for the next $p - 1$ following steps, then agent $A_i \notin LIVELOCK_{p-1}$.
The proof is straightforward.

According with the precedent example, defined $p = 3$, there exist two different paths $p'_{A_i} = \{1, 4, 7\}$ e $p''_{A_i} = \{1, 5, 9\}$ in $P_{safe}$ that the agent can follow without being in livelock in 3 steps.

## VI. SIMULATION OF THE COORDINATION ALGORITHM

Many tests have been done proving the correctness and the efficiency of the coordination algorithm proposed. In a first example four agents all requesting the same resource have been considered as represented in Fig. 14: In particular, desired paths are $path_1 = \{55, 56, 57\}$, $path_2 = \{66, 56, 46\}$, $path_3 = \{57, 56, 55\}$, $path_4 = \{46, 56, 66\}$ and $Pr_i = i$ for $i = 1, \ldots, 4$. Such configuration is a deadlock one in case the replanning algorithm is not taken into account.

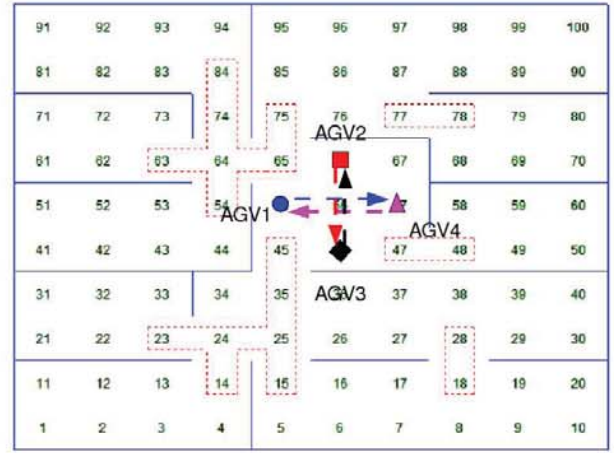

Fig. 14. Example of a deadlock configuration in case of no-replanning

On the contrary, according to the rule defined in the shared coordination protocol, agents $A_1$ e $A_2$, with minor priority, loose competitions for the *frontal* macro resources with agents $A_3$ and $A_4$, respectively. As a consequence, they replan their paths toward their goal nodes without crossing the central node 56: $path_1 = \{55, 45, 46, 45, 57\}$, $path_2 = \{66, 65, 55, 45, 46\}$. Moreover, agent $A_3$ loses the competition for the *cross* macro resource with agent $A_4$ and waits its passage before taking the central node. The result of the coordination protocol is represented in Fig. 15.

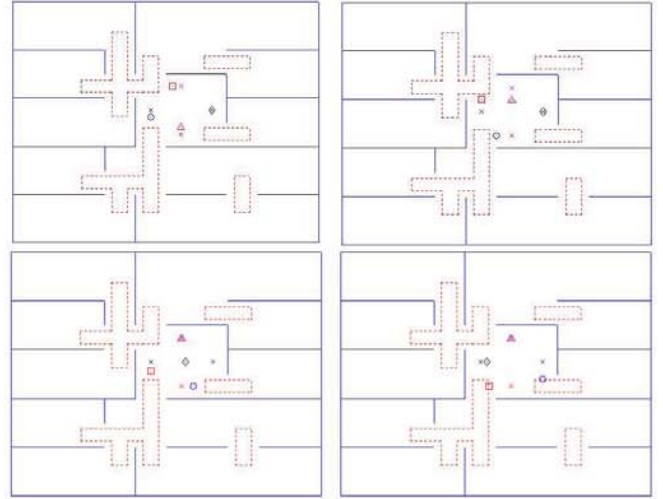

Fig. 15. Example of deadlock resolution through re-planning

Another simulation in a deadlock configuration (if replanning is not taken into account) is reported in Fig. 16.

Videos of simulations can be found at [26].

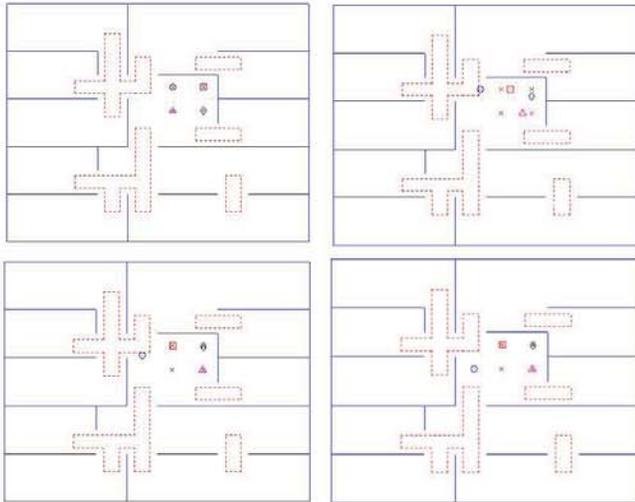

Fig. 16.   Example of deadlock resolution through re-planning

## VII. CONCLUSIONS AND FUTURE WORK

A fully distributed control algorithm for the motion coordination of autonomous guided vehicles for industrial application has been presented. The proposed algorithm is characterized by the following properties:

- Distributed: shared coordination protocol between agents based on rules for the access to shared resources;
- Local: resources are negotiated locally based on the information of agents in a given communication radius;
- Scalable: the computational cost does not increase with the number of agents in the environment.

Mutual access to resources has been proved for the proposed approach. Condition on the maximum number of AGVs is given to ensure the absence of deadlocks during system evolutions. Finally conditions to verify a local livelocks have also been proposed.
Future works could improve efficiency of the proposed algorithm, or take into account on-line deadlock/livelock identification so that agents can replan before the stall.

## VIII. ACKNOWLEDGMENTS

Authors wish to thank Davide Di Baccio for his collaboration in improving a previous version of this paper.